%% file: main.tex
\documentclass{article}

\PassOptionsToPackage{numbers,compress}{natbib}

\usepackage[eandd,final]{neurips_2026}

\usepackage[utf8]{inputenc}
\usepackage[T1]{fontenc}

\usepackage{hyperref}
\usepackage{url}
\usepackage{graphicx}
\usepackage{booktabs}
\usepackage{amsmath}
\usepackage{amsfonts}
\usepackage{nicefrac}
\usepackage{microtype}
\usepackage{xcolor}

\newcommand{\editcolor}{black}
\DeclareRobustCommand{\edited}[1]{{\color{\editcolor}#1}}
\newcommand{\reviewcolor}{black}
\DeclareRobustCommand{\newedit}[1]{{\color{\reviewcolor}#1}}
\edef\restoretemplatespacing{%
  \textfloatsep=\the\textfloatsep\relax
  \intextsep=\the\intextsep\relax
  \floatsep=\the\floatsep\relax
  \abovecaptionskip=\the\abovecaptionskip\relax
  \belowcaptionskip=\the\belowcaptionskip\relax
  \tabcolsep=\the\tabcolsep\relax}
\input{preamble}
\restoretemplatespacing

\title{EEGFaceSem: An EEG Benchmark with Paired Generative Latents for Semantic Visual Modeling}

\author{%
  Jun Ma \\
  University of Helsinki \\
  Helsinki, Finland \\
  \texttt{jun.ma@helsinki.fi}
  \And
  Michiel M. Spap\'e \\
  University of Macau \\
  Macau, China \\
  \texttt{mspape@um.edu.mo}
  \AND
  Keith M. Davis III \\
  LUT University \\
  Lahti, Finland \\
  \texttt{keith.davis@lut.fi}
  \And
  Tuukka Ruotsalo \\
  University of Copenhagen \\
  Copenhagen, Denmark \\
  LUT University \\
  Lahti, Finland \\
  \texttt{tr@di.ku.dk}
}

\begin{document}
\maketitle

\begin{abstract}
Electroencephalography (EEG) is widely used to measure brain activity for brain-computer interface applications for its non-invasive nature and high temporal resolution.
However, progress in modeling the neural responses to visual semantics has been limited by a reliance on \edited{fixed image collections} that lack the fine-grained control needed to isolate fine-grained changes in visual features and their relation to perception.
In this work, we present EEGFaceSem, a large-scale dataset pairing high-quality EEG with photorealistic face images synthesized from a generative model.
We provide the generative latent vector for each stimulus, enabling research into the continuous mapping between a visual latent space and neural representations.
Each participant's brain responses were recorded while they attended to specific semantic features (e.g., smiling, hair color) in a rapid serial visual presentation paradigm.
The dataset contains EEG signals from 30 participants, totaling 67,200 time-locked EEG recordings.
We present benchmarks for single-trial semantic classification under single-subject, cross-subject, and subject-adapted settings, \edit{as well as for cross-subject voting that aggregates per-stimulus predictions across participants}.
We further demonstrate the dataset's unique utility \edit{with two complementary applications: deriving semantic directions in the latent space for brain-supervised image editing, and conditioning a generative model on brain-decoded labels for brain-supervised image generation.}
Our dataset paves the way for advancing brain-computer interfaces (BCI) for semantic-level visual saliency detection and brain-supervised generative modeling.
Our dataset and code are openly released at OSF (\url{https://osf.io/2guk3/}) and at HuggingFace \url{https://huggingface.co/datasets/yefllower/EEGFaceSem}.
\end{abstract}

\begin{figure*}[t!]
  \centering
  \includegraphics[width=\textwidth]{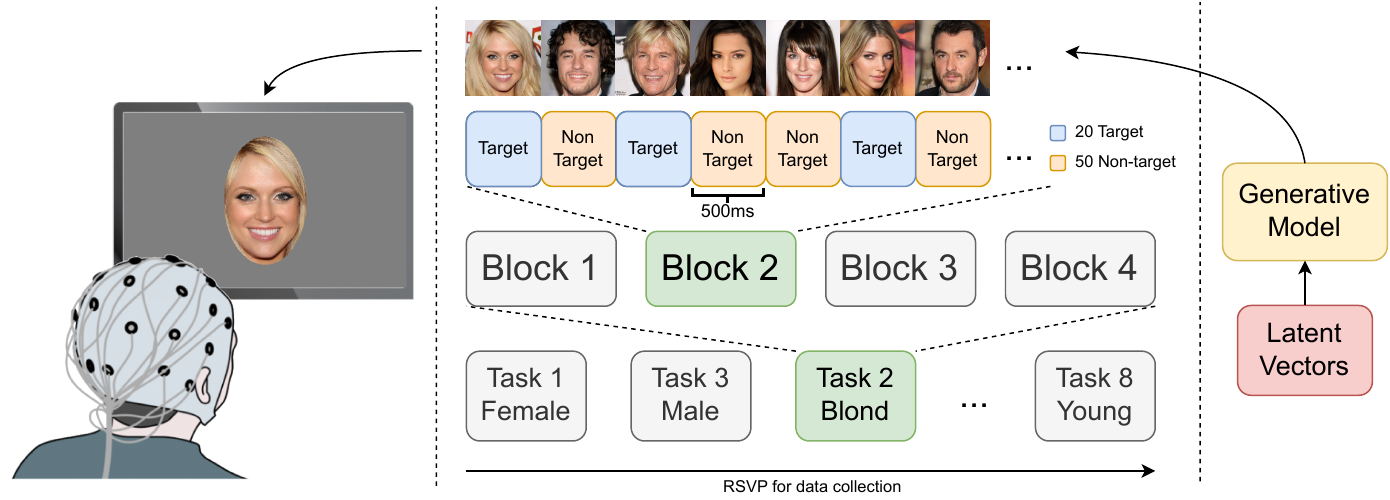}
  \caption{Overview of the data acquisition pipeline, from generative stimulus creation to neural recording.
  Central to our paradigm, visual stimuli are synthesized by passing latent vectors through a pre-trained generative model.
  During the experiment, a participant views a sequence of these generated faces presented in a rapid serial visual presentation (RSVP) stream.
  For each task (e.g., 'smiling'), each stimulus is contextually defined as a 'target' or 'non-target', as shown by the corresponding label sequence. This procedure is repeated across multiple blocks for each of the eight semantic tasks, allowing for the collection of a rich set of time-locked EEG responses.}
  \label{fig:overview}
\end{figure*}

\section{Introduction}
Understanding how the human brain perceives and attends to salient visual information is a fundamental challenge in cognitive neuroscience \cite{desimone1995neural} and computer vision \cite{itti2005models}.
A primary bottleneck in this research has been the nature of the visual stimuli.
\edited{It can be difficult to} disentangle the neural correlates of specific semantic features (e.g., 'smiling') from confounding low-level visual properties (e.g., lighting, background clutter).
Furthermore, a fixed collection of images only provides sparse samples from the vast, continuous space of possible visual stimuli, hindering the study of how neural representations evolve along meaningful semantic dimensions.

To address these limitations, we introduce EEGFaceSem, a large-scale EEG dataset that pioneers a new paradigm for stimulus design by leveraging the latent space of a generative model.
\edited{Our stimuli are} photorealistic face images synthesized by a progressive GAN~\cite{karras2018progressive} pre-trained on the CelebA dataset~\cite{liu2015faceattributes}. \edited{The presented images are a finite sample from the generator's continuous latent space.}
This approach offers two distinct advantages.
First, it enables unprecedented control over semantic attributes while maintaining visual richness.
Second, and most critically, we provide not only the images and EEG recordings, but also the 512-dimensional latent vectors used to generate each stimulus.
This direct link between the neural data and the underlying generative latent space transforms the dataset from a simple collection of image-EEG pairs into a powerful tool for modeling the continuous, high-dimensional mapping between brain activity and visual semantics.
It opens the door to novel research in brain-supervised generative modeling\newedit{~\cite{cr_kangassalo2020neuroadaptive,cr_torre2020relevance}}, the exploration of joint EEG-latent embedding spaces, and the ability to synthesize novel stimuli guided by implicit neural feedback\newedit{~\cite{cr_torre2024perceptual,cr_grizou2025selfcalibrating}}.

\edited{Here, ``semantic'' refers to task-defined attributes shared across different faces. The recorded task-evoked responses include attention and P300 components.}

We provide detailed benchmarks for single-trial semantic classification under single-subject, cross-subject, and subject-adapted settings, as well as for an aggregated multi-trial evaluation.
We further demonstrate the dataset's unique utility with a exemplary application, using neural signals to derive semantic directions in the latent space for brain-supervised image editing~\cite{davis2022brain}.

\section{Neurophysiological datasets for semantic visual processing}
The landscape of neurophysiological datasets for vision research is characterized by a fundamental trade-off between stimulus complexity and experimental control.
On one end of the spectrum, datasets employing simple, symbolic stimuli, such as the P300 speller in BNCI2015\_003 \cite{guger2009many}, offer precise control but lack the visual richness required to study complex semantic processing.
On the other end, datasets featuring large collections of natural images, such as THINGS-EEG \cite{grootswagers2022human}, Visual Object Recognition EEG \cite{gifford2022large} and EEG-ImageNet \cite{zhu2024eeg}, provide rich visual content.
However, these stimuli represent discrete, uncontrolled samples from the visual world, making it difficult to disentangle neural responses to high-level semantics from the confounding effects of low-level visual features like luminance and spatial frequency \cite{rossion2014understanding}.

Our dataset is designed to bridge this gap.
By providing time-locked EEG responses to a large set of synthetically generated face images, we combine rich semantic diversity with fine-grained computational control.
This approach allows for the systematic manipulation of high-level semantic attributes (e.g., age, emotion) while mitigating the confounds of uncontrolled low-level features.
As human faces are a well-established stimulus for eliciting strong neural responses in multi-modal datasets \cite{wakeman2015multi}, our work is the first to leverage controlled stimuli with a generative latent space to specifically investigate semantic visual processing.
\edited{Other resources include EEG2Video for dynamic reconstruction~\cite{cr_liu2024eeg2video}, Alljoined1 for EEG-to-image decoding~\cite{cr_xu2024alljoined}, EEG-Face for face-stimulus EEG~\cite{cr_zhang2025eegface}, and the Natural Scenes Dataset and THINGS-MEG for fMRI and MEG~\cite{cr_allen2022nsd,cr_heba2022things}.}
This distinguishes our contribution from other recent efforts, such as those focusing on mental imagery rather than controlled visual stimuli \cite{rybavr2025simultaneous}.
Table~\ref{table:datasets} provides a detailed comparison of our dataset with prior datasets.

\begin{table*}[h]
    \begin{center}
    \setlength{\tabcolsep}{3pt}
    \caption{Comparison of EEG Datasets. }
    \label{table:datasets}
    \vspace{0.2cm}
    \resizebox{\linewidth}{!}{
    \begin{tabular}{ l|cccccc }
    \toprule
    Dataset         & \#Subject & \edited{Visual stimulus content} & Standardized  & Semantic & Latent Representation & Modality \\
    \midrule
    Grootswagers, Tijl, et al. \cite{grootswagers2022human}
                    & 50        &   \edited{Object images}               &      no       &    yes   &  no &  EEG     \\
    Guger, Christoph, et al. \cite{guger2009many}
                    & 10        &   \edited{Characters}                &      yes      &    yes   &  no &  EEG     \\
    Gifford, Alessandro T., et al. \cite{gifford2022large}
                    & 10        &   \edited{Object images}               &      no       &    yes   &  no &  EEG     \\
    Wakeman, Daniel G., and Richard N. Henson. \cite{wakeman2015multi}
                    & 19        &   \edited{Face images}               &      no       &    yes   &  no &  EEG+MEG+fMRI     \\
    Ryb{\'a}{\v{r}}, Milan, Riccardo Poli, and Ian Daly. \cite{rybavr2025simultaneous}
                    & 12        &   \edited{Object images}               &      no       &    no    &  no &  EEG+fNIRS     \\
    \midrule
    Ours
                    & 30        &   \edited{Generated faces}               &      yes      &    yes   & yes &  EEG     \\
    \bottomrule
    \end{tabular}
    }
    \end{center}
\end{table*}

\begin{table*}[h]
  \centering
  \caption{Summary of dataset statistics before and after pre-processing.}
  \resizebox{\linewidth}{!}{
  \begin{tabular}{cccccccc}
  \hline
  subject & channel &  task &  blocks per task &  target per block &  non-target per block &  sampling rate & total trials\\
  30 & 32 & 8 & 4 & 20 & 50 & 1000 & 67200\\
  \hline
  female & male & blond & darkhaired & smiles & nosmile & old & young \\
  7460 & 7416 & 7354 & 7248 & 7365 & 7553 & 7493 & 7459 \\
  \hline
  \end{tabular}
  }
  \label{tab:summary}
\end{table*}

\section{EEG data acquisition}

In this section we explain in details the participants, stimuli generation, experimental protocol, and preprocessing procedures used to collect and validate the EEG signals included in the dataset.

\subsection{Participants}
Thirty-one healthy volunteers were recruited; one withdrew before completion, yielding data from $N{=}30$ participants (self-reported: 13 female, 17 male; mean age 28, SD=7.14, range 18-45), all with normal or corrected-to-normal vision and no history of neurological disease.
Informed consent was obtained from all participants in accordance with the ethical principles for medical research involving human subjects outlined in the Declaration of Helsinki.
Participants were compensated for their time with vouchers for local cinema.
The study was approved by the ethical committee of \edited{the University of Helsinki}.

\subsection{Stimuli Generation and Grouping}
We used synthetically generated, photorealistic face images rather than real images of identifiable individuals. This design \edited{reduces familiarity with real identities and supports selection of semantic attributes}.

A total of 70,000 face images were generated using a progressive GAN~\cite{karras2018progressive} pre-trained on the CelebA dataset~\cite{liu2015faceattributes}.
The progressive GAN was sampled by drawing from a 512-dimensional multivariate normal distribution. \edited{PGAN provided a fixed 512-dimensional input for interpolation and attribute editing; modern high-fidelity diffusion face generators were unavailable when the data were collected.}
All generated images were manually screened by \edited{one researcher} to remove those with visual artifacts, unrealistic features, or obvious signs of artificiality.
The remaining images were categorized into eight semantic groups based on visual attributes: smiling, not smiling, female, male, young, old, dark hair, and blond hair.
To minimize background confounds, an elliptic grey frame was overlaid on all images to mask non-facial regions.

\edit{To verify that the screened stimuli express the intended semantic contrasts, we ran an external attribute audit using CLIP ViT-L/14~\cite{radford2021clip} in a zero-shot configuration. Across the four target contrasts (gender, smile, hair colour, age), the classifier separates the screened positive and negative groups with AUC $\geq 0.816$ (max $0.997$); the residual pairwise phi-coefficient coupling introduced by the screening is below $0.015$ in absolute value relative to a random reference batch sampled from the same generator. \edited{Per-contrast separation is reported in Table~\ref{tab:attribute_audit_separation}; residual associations are in Appendix~\ref{appendix:attribute_audit}.}}

\begin{table}[!htbp]
  \centering
  \small
  \edit{\caption{External attribute audit: per-contrast separation of the screened stimuli under CLIP ViT-L/14 zero-shot probing. Each row reports the CLIP softmax probability of the positive attribute on the target and non-target image sets, together with the AUC.}\label{tab:attribute_audit_separation}}
  \begin{tabular}{lcccc}
    \toprule
    Contrast & Score & Positive group & Negative group & AUC \\
    \midrule
    Female vs Male & $P(\mathrm{Male})$ & $0.948 \pm 0.163$ & $0.016 \pm 0.094$ & $0.997$ \\
    NoSmile vs Smiling & $P(\mathrm{Smiling})$ & $0.617 \pm 0.241$ & $0.189 \pm 0.218$ & $0.883$ \\
    Dark vs Blond hair & $P(\mathrm{Blond})-P(\mathrm{Dark})$ & $0.219 \pm 0.268$ & $-0.769 \pm 0.175$ & $0.995$ \\
    Old vs Young & $P(\mathrm{Young})$ & $0.894 \pm 0.123$ & $0.691 \pm 0.226$ & $0.816$ \\
    \bottomrule
  \end{tabular}
\end{table}

\subsection{Apparatus and environment}
EEG was recorded using 32 Ag/AgCl electrodes placed according to the international 10-20 system \cite{jasper1958ten}, connected to a QuickAmp USB amplifier (BrainProducts GmbH, Gilching, Germany) \edit{with a sampling rate of 1,000 Hz}.
Two pairs of bipolar electrodes were used to monitor eye movements and blinks (1 cm lateral to the left and right canthi, and 2 cm above and below the right pupil).
Stimuli were presented on a screen at a fixed viewing distance in a quiet room.

\subsection{Task and block design}
\label{sec:block_design}
The experiment was structured into eight distinct tasks, each corresponding to one semantic attribute (e.g., smiling, blond hair), as shown in Figure~\ref{fig:overview}.
Each task was composed of four repeated blocks to ensure sufficient data quality, resulting in a total of 32 blocks per participant.
Before the first block of each new task, participants completed a demonstration (pre-task) phase.
In this phase, four example images were shown, and participants were asked to manually select those matching the target attribute.
This ensured participants understood the semantic feature for the task.
The demonstration images were not used in the main experiment.

\edited{Each block fixes the attended attribute and contains both target and non-target images. Each task comprises four consecutive blocks; task-group order varies across participants. Image order is randomized within blocks. All reported classifiers distinguish target from non-target within one task. Temporal dependence and block-state effects remain possible~\cite{li2020perils}.}

\subsection{Experimental procedure}
Within each of the 32 blocks, participants viewed a rapid serial visual presentation (RSVP) stream of 70 unique images (20 target, 50 non-target) in a random order.
Each image was displayed for 500 ms, with no pause between images. \edited{Epochs are onset-aligned windows from continuous RSVP and can contain responses to adjacent images.}
Participants were instructed to passively observe the images and mentally note whenever an image matched the task's target attribute.
No physical or behavioral response was required.
This design resulted in a total of 2,240 trials (8 tasks $\times$ 4 blocks $\times$ 70 trials) per participant for the main experiment.

\subsection{Data preprocessing, annotation, and analysis}
\paragraph{EEG Preprocessing}
After the data acquisition, standard signal cleaning procedures were employed~\cite{luck_artifact_2014} to improve the signal-to-noise ratio.
These were restricted to only automatic procedures that do not require any additional labeled data, including a band-pass filter in the frequency range 0.2-35 Hz and time-locking to \textit{epochs} ranging from -200 to 900 ms relative to stimulus onset with baseline correction based on a pre-stimulus period of -200 to 0 ms.
Eyeblink artifacts were removed using a threshold-based heuristic.
An average of 1978 epochs per participant remained after pre-processing, as shown in Table~\ref{tab:summary}.

\begin{figure*}[h]
    \centering
    \includegraphics[width=0.62\linewidth]{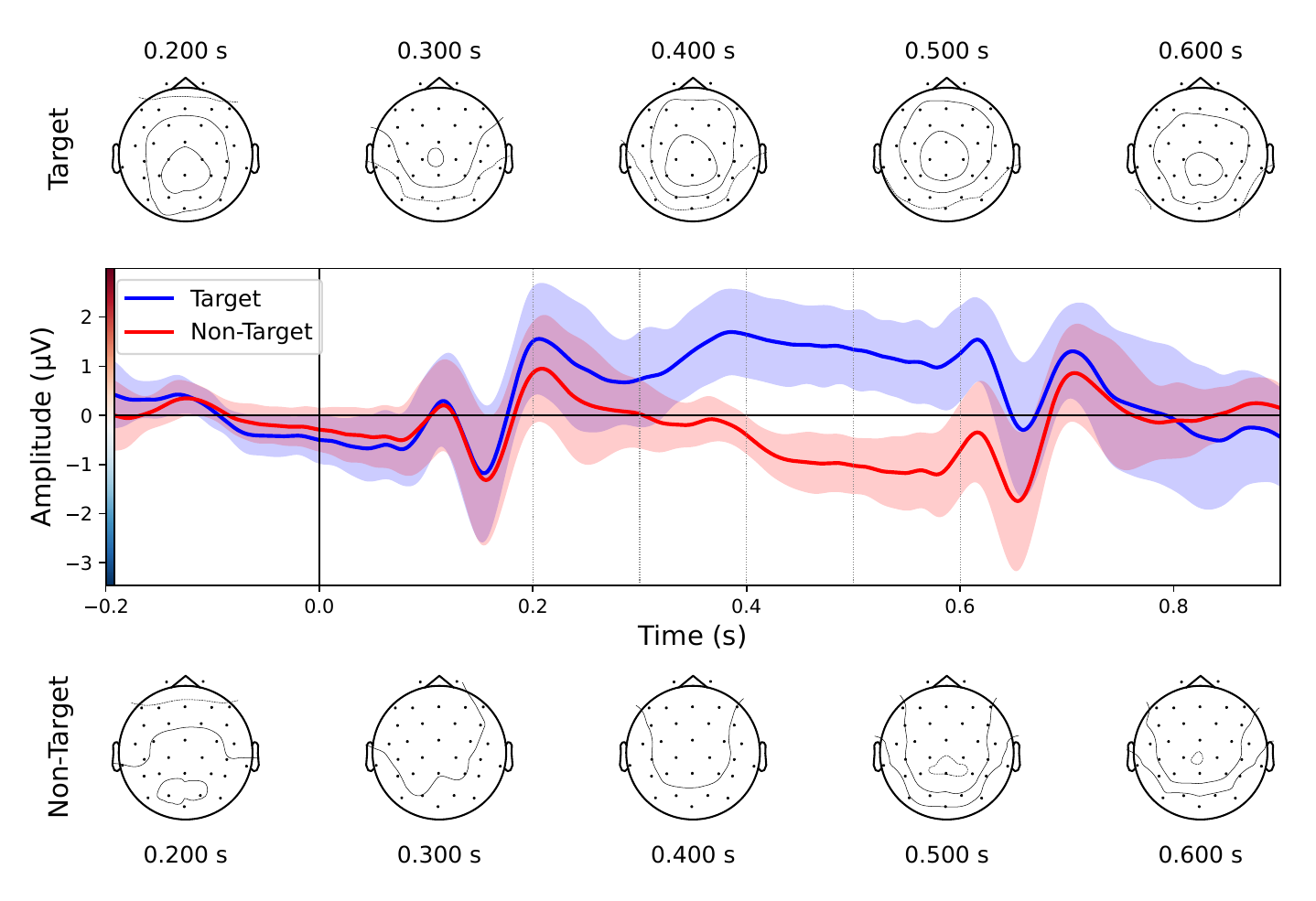}
    \includegraphics[width=0.37\linewidth]{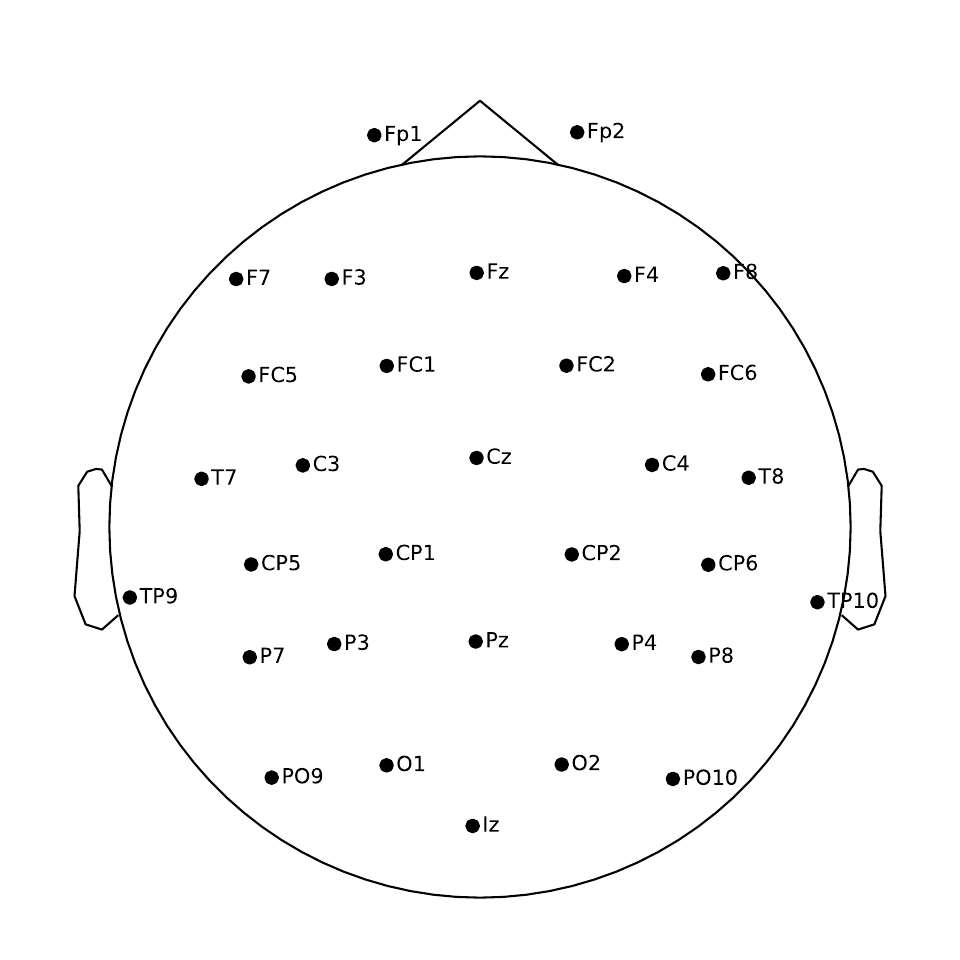}
  \vspace{-0.2cm}
  \caption{\textbf{Left}: Average event-related potentials (ERPs) across participants at the Pz electrode for target and non-target stimuli, illustrating a P300 effect. Topoplots show the scalp distribution of the ERP at specific time points (0.2s, 0.3s, 0.4s, 0.5s, 0.6s). \textbf{Right}: Electrode positions on the scalp.}
  \label{fig:erp}
\end{figure*}

\paragraph{Ground truth annotation}
Each stimulus was labeled as target or non-target for the current task.
The dataset includes both raw and preprocessed EEG, stimulus metadata, and semantic labels.

\paragraph{ERP analysis}
Event-related potentials (ERPs) \cite{chong2023comparing,hoffman1983event}, such as the P300 component \cite{picton1992p300}, were analyzed to validate cognitive responses to target stimuli, as shown in Figure~\ref{fig:erp}.
The data averaged over participants shows a typical P300-effect with amplified potentials for target stimuli starting at 250 ms after stimuli onset and lasting until 600 ms.
This finding confirms that, on population average, the experiment resulted in an expected ERP effect.

\section{Benchmark Machine Learning Experiments}
\label{sec:experiments}

To establish a comprehensive performance baseline for our dataset and to probe its utility for brain-computer interface (BCI) and cognitive neuroscience research, we define a series of benchmark classification experiments.
A primary challenge in this dataset is the context-dependent nature of the ground-truth labels;
for example, a binary "target" response for a "blond hair" task represents a distinct cognitive event from a "target" response for another task such as "dark hair".
A naive mixing of all data would confound these distinct neural processes. Our experimental design therefore explicitly addresses this problem by defining \edited{eight} separate classification tasks.
These tasks are evaluated using three distinct training strategies to test for generalizability and single-subject specialization.

\begin{table*}[t]
  \centering
  \caption{Performance of the single-subject classifier on all tasks, evaluated using 5-fold cross-validation with explicit labels. Values are reported as the mean AUC $\pm$ standard deviation calculated across all test folds.}
  \label{tab:classifier_auc_single}
  \resizebox{\linewidth}{!}{\begin{tabular}{r|ccccc}
\toprule
Task & LDA & LR & MLP & EEGNet & EEGPT \\
\midrule
Female          & $0.7895 \pm 0.0820$ & $0.7929 \pm 0.0787$ & $0.7382 \pm 0.0738$ & $0.8239 \pm 0.0785$ & $0.7048 \pm 0.0872$ \\
Male            & $0.7248 \pm 0.0852$ & $0.7351 \pm 0.0852$ & $0.6652 \pm 0.0795$ & $0.7374 \pm 0.1017$ & $0.6105 \pm 0.0946$ \\
Blond           & $0.7741 \pm 0.0924$ & $0.7816 \pm 0.0880$ & $0.7026 \pm 0.0926$ & $0.7853 \pm 0.0935$ & $0.6523 \pm 0.1024$ \\
Darkhaired      & $0.7414 \pm 0.0767$ & $0.7484 \pm 0.0764$ & $0.6652 \pm 0.0672$ & $0.7506 \pm 0.0856$ & $0.6424 \pm 0.0939$ \\
Smiles          & $0.7271 \pm 0.0842$ & $0.7379 \pm 0.0850$ & $0.6705 \pm 0.0914$ & $0.7648 \pm 0.0873$ & $0.6138 \pm 0.0871$ \\
Nosmile         & $0.7058 \pm 0.0905$ & $0.7114 \pm 0.0973$ & $0.6546 \pm 0.0837$ & $0.7384 \pm 0.0830$ & $0.6040 \pm 0.0969$ \\
Old             & $0.6496 \pm 0.0789$ & $0.6562 \pm 0.0806$ & $0.6251 \pm 0.0742$ & $0.6526 \pm 0.0833$ & $0.5766 \pm 0.0675$ \\
Young           & $0.6524 \pm 0.0863$ & $0.6608 \pm 0.0816$ & $0.6188 \pm 0.0786$ & $0.6798 \pm 0.0994$ & $0.5856 \pm 0.0727$ \\
\midrule
Mean            & $0.7206 \pm 0.0475$ & $0.7280 \pm 0.0469$ & $0.6675 \pm 0.0364$ & $0.7416 \pm 0.0513$ & $0.6238 \pm 0.0389$ \\
\bottomrule
  \end{tabular}}
\end{table*}

\subsection{Training and Evaluation Strategies}
\label{sec:strategies}

We employ three strategies to train and evaluate classifier models, adapted from common methodologies in EEG literature \cite{kunjan2021necessity,lotte2007review,azab2018review}.

\paragraph{Cross-Subject.}
This strategy, often called Leave-One-Subject-Out (LOSO) cross-validation, evaluates a model's ability to generalize to a completely unseen participant \cite{kunjan2021necessity}.
For $N=30$ subjects, the dataset is split into $N$ folds, where each fold contains all data from a single subject. The model is trained on data from $N-1$ subjects and tested on the held-out subject.
This process is repeated $N$ times, with performance averaged across all folds.
From the training set, one additional subject is held out for validation, and the remaining $N-2$ subjects are used for training.

\paragraph{Single-Subject.}
This strategy evaluates a model trained and tested exclusively on a single participant's data, testing the feasibility of creating a highly personalized model.
For each of the $N$ subjects, we perform a $k$-fold cross-validation using only that subject's data.
The final metric is the average performance across all $N$ subjects and all $k$ folds.

\paragraph{Subject-Adapted.}
This strategy simulates a practical BCI scenario where a general model is personalized with a small amount of user data.
We first train a cross-subject model on data from $N-1$ subjects.
We then fine-tune this pre-trained model on a subset of the target subject's data and evaluate it on the remaining unseen data from that same subject with $k$-fold cross-validation.
The average performance across all $N$ subjects and all $k$ folds is reported.

\subsection{Classifier Models}
\label{sec:classifiers}

We select a representative collection of classifiers to provide a robust benchmark, encompassing both classical machine learning and modern deep learning approaches.

\paragraph{Baseline Classifiers.}
We include several strong and widely-used baselines for EEG classification:
Linear Discriminant Analysis (LDA)~\cite{blankertz2011single}, a robust method for finding a discriminative subspace;
Logistic Regression (LR) with L2 regularization \cite{tomioka2006logistic};
a simple Multi-Layer Perceptron (MLP);
EEGNet \cite{lawhern2018eegnet}, a compact convolutional neural network (CNN) specifically designed for EEG signals;
and EEGPT~\cite{wang2024eegpt}, a recent transformer-based foundation model for EEG.
For EEGPT, we follow the original paper and use a linear head on top of the frozen pre-trained EEGPT model, and a scaling layer before the patch embedding layer.

\subsection{EEG Data Representation}
\label{sec:representation}
For all benchmark classification models, the input consists of preprocessed EEG epochs.
Each epoch is a matrix of channels $\times$ time points, corresponding to a time window from -200ms to +900ms relative to the image stimulus onset.
With a sampling rate of 1000Hz and 32 channels, this results in a $32 \times 1101$ input matrix.
This window is chosen to capture pre-stimulus period, early visual-evoked potentials and later cognitive components associated with the decision-making process.
This direct representation allows the models to learn discriminative spatio-temporal patterns from the signal itself.

\begin{table*}[t]
    \centering
    \caption{Performance of the cross-subject classifier on all tasks, evaluated using \edited{leave-one-subject-out cross-validation} with explicit labels. Values are reported as the mean AUC $\pm$ standard deviation calculated across all test folds.}
    \label{tab:classifier_auc_cross}
    \resizebox{\linewidth}{!}{\begin{tabular}{r|ccccc}
\toprule
Task & LDA & LR & MLP & EEGNet & EEGPT \\
\midrule
Female          & $0.6191 \pm 0.0627$ & $0.8039 \pm 0.0649$ & $0.8120 \pm 0.0634$ & $0.8338 \pm 0.0684$ & $0.8022 \pm 0.0602$ \\
Male            & $0.5869 \pm 0.0573$ & $0.7221 \pm 0.0637$ & $0.7200 \pm 0.0737$ & $0.7632 \pm 0.0689$ & $0.6806 \pm 0.0561$ \\
Blond           & $0.6301 \pm 0.0771$ & $0.7759 \pm 0.0660$ & $0.7798 \pm 0.0693$ & $0.8075 \pm 0.0691$ & $0.7487 \pm 0.0693$ \\
Darkhaired      & $0.6130 \pm 0.0664$ & $0.7411 \pm 0.0779$ & $0.7429 \pm 0.0818$ & $0.7699 \pm 0.0784$ & $0.7090 \pm 0.0738$ \\
Smiles          & $0.6077 \pm 0.0643$ & $0.7678 \pm 0.0584$ & $0.7691 \pm 0.0612$ & $0.8021 \pm 0.0650$ & $0.7217 \pm 0.0561$ \\
Nosmile         & $0.5820 \pm 0.0641$ & $0.7207 \pm 0.0889$ & $0.7220 \pm 0.0823$ & $0.7599 \pm 0.0874$ & $0.6941 \pm 0.0642$ \\
Old             & $0.5813 \pm 0.0548$ & $0.6724 \pm 0.0686$ & $0.6613 \pm 0.0658$ & $0.6906 \pm 0.0840$ & $0.5890 \pm 0.0563$ \\
Young           & $0.6063 \pm 0.0508$ & $0.7134 \pm 0.0634$ & $0.7065 \pm 0.0689$ & $0.7452 \pm 0.0718$ & $0.6572 \pm 0.0522$ \\
\midrule
Mean            & $0.6033 \pm 0.0169$ & $0.7397 \pm 0.0389$ & $0.7392 \pm 0.0442$ & $0.7715 \pm 0.0412$ & $0.7003 \pm 0.0591$ \\
\bottomrule
    \end{tabular}}
\end{table*}

\subsection{Benchmark Tasks and Results}
\label{sec:tasks_and_results}

We evaluate all classifiers on the binary classification task (target vs. non-target), with performance metrics averaged across all eight semantic tasks.
The area under the ROC Curve (AUC) is used as the primary metric.

\paragraph{Single-Trial Performance.}
The core of our benchmark is the single-trial classification performance, where each EEG epoch is decoded independently.
Tables \ref{tab:classifier_auc_single}, \ref{tab:classifier_auc_cross}, and \ref{tab:classifier_auc_adapt} present these results for the single-subject, cross-subject, and subject-adapted strategies, respectively.

The cross-subject results in Table \ref{tab:classifier_auc_cross} show that all models perform well above chance, with EEGNet achieving the highest mean AUC of $0.7715 \pm 0.0412$.
This establishes that a generalizable neural signal for the tasks exists across subjects and tasks.
As shown in Table \ref{tab:classifier_auc_single}, performance generally drops in the single-subject setting except for LDA, confirming that models benefit from training on a large dataset.
The subject-adapted strategy results in Table \ref{tab:classifier_auc_adapt} demonstrates the best single-trial performance for MLP and EEGNet, with EEGNet reaching an AUC of $0.8063 \pm 0.0427$.
This demonstrates the effectiveness of fine-tuning a general model for personalization, compared to the other two strategies.
Across these settings, the transformer-based EEGPT does not outperform the simpler MLP and EEGNet, potentially due to its large parameter count requiring more data for effective adaptation, or the gap between its pre-training data and our task.

\paragraph{\edit{Cross-Subject Voting.}}
\edit{Many BCI systems improve accuracy by aggregating responses over multiple sources -- multiple trials within a subject \cite{lopez2022state,giroldini2016new,hajra2021event,hu2010novel}, or, when the same stimulus is shown to many subjects, across subjects.
Because each stimulus in our dataset is shown to all $N{=}30$ participants exactly once per session, we evaluate the latter setting: for every stimulus image, we collect the single-trial cross-subject (LOSO) predictions from all participants who viewed it and combine them by majority voting. The confidence of the voted prediction is the average probability of the winning class across the contributing trials.}

The results of this aggregation are presented in Table \ref{tab:classifier_acc}. The performance gains are substantial, with EEGNet reaching an AUC of $0.9539 \pm 0.0287$.
This significant improvement indicates that while individual trials contain noise and individual variations, the underlying cognitive response signal is highly robust and consistent across the population.

\begin{table*}[t]
  \centering
  \caption{Performance of the subject-adapted classifier on all tasks, evaluated using 5-fold cross-validation with explicit labels. Values are reported as the mean AUC $\pm$ standard deviation calculated across all test folds. Models that do not support finetuning are not included.}
  \label{tab:classifier_auc_adapt}
\begin{tabular}{r|ccc}
\toprule
Task & MLP & EEGNet & EEGPT \\
\midrule
Female          & $0.8261 \pm 0.0932$ & $0.8643 \pm 0.0781$ & $0.7981 \pm 0.0818$ \\
Male            & $0.7463 \pm 0.1015$ & $0.8171 \pm 0.0951$ & $0.6501 \pm 0.0640$ \\
Blond           & $0.7935 \pm 0.1015$ & $0.8427 \pm 0.0915$ & $0.7315 \pm 0.0892$ \\
Darkhaired      & $0.7697 \pm 0.1065$ & $0.8219 \pm 0.0973$ & $0.6703 \pm 0.0865$ \\
Smiles          & $0.7728 \pm 0.0980$ & $0.8233 \pm 0.0949$ & $0.6990 \pm 0.0602$ \\
Nosmile         & $0.7444 \pm 0.1027$ & $0.8012 \pm 0.1044$ & $0.6732 \pm 0.0580$ \\
Old             & $0.6757 \pm 0.1078$ & $0.7259 \pm 0.1016$ & $0.5833 \pm 0.0675$ \\
Young           & $0.7080 \pm 0.1067$ & $0.7537 \pm 0.1073$ & $0.6539 \pm 0.0733$ \\
\midrule
Mean            & $0.7546 \pm 0.0444$ & $0.8063 \pm 0.0427$ & $0.6824 \pm 0.0592$ \\
\bottomrule
\end{tabular}
\end{table*}

\begin{table*}[t]
  \centering
  \edit{\caption{Performance of cross-subject voting under the LOSO strategy: per-stimulus predictions from all participants are combined via majority voting. The reported AUC is computed after combining individual model predictions. The standard deviation is estimated using 1,000 bootstrap resamples.}\label{tab:classifier_acc}}
  \resizebox{\linewidth}{!}{\begin{tabular}{r|ccccc}
  \toprule
  Task & LDA & LR & MLP & EEGNet & EEGPT \\
  \midrule
  Female               & 0.7960 $\pm$ 0.0228& 0.9736 $\pm$ 0.0084& 0.9801 $\pm$ 0.0065& 0.9830 $\pm$ 0.0058& 0.9720 $\pm$ 0.0079 \\
  Male                 & 0.7137 $\pm$ 0.0257& 0.9543 $\pm$ 0.0118& 0.9605 $\pm$ 0.0091& 0.9695 $\pm$ 0.0100& 0.9328 $\pm$ 0.0126 \\
  Blond                & 0.8641 $\pm$ 0.0205& 0.9862 $\pm$ 0.0073& 0.9921 $\pm$ 0.0043& 0.9881 $\pm$ 0.0069& 0.9765 $\pm$ 0.0095 \\
  Darkhaired           & 0.7795 $\pm$ 0.0262& 0.9644 $\pm$ 0.0107& 0.9654 $\pm$ 0.0103& 0.9668 $\pm$ 0.0101& 0.9407 $\pm$ 0.0140 \\
  Smiles               & 0.7386 $\pm$ 0.0208& 0.9253 $\pm$ 0.0117& 0.9442 $\pm$ 0.0102& 0.9608 $\pm$ 0.0084& 0.8942 $\pm$ 0.0137 \\
  Nosmile              & 0.6811 $\pm$ 0.0206& 0.9104 $\pm$ 0.0130& 0.9115 $\pm$ 0.0123& 0.9334 $\pm$ 0.0112& 0.8696 $\pm$ 0.0155 \\
  Old                  & 0.6802 $\pm$ 0.0228& 0.8543 $\pm$ 0.0166& 0.8591 $\pm$ 0.0162& 0.8967 $\pm$ 0.0136& 0.7421 $\pm$ 0.0215 \\
  Young                & 0.7412 $\pm$ 0.0213& 0.9057 $\pm$ 0.0134& 0.8898 $\pm$ 0.0144& 0.9329 $\pm$ 0.0107& 0.8460 $\pm$ 0.0167 \\
  \midrule
  Mean                 & 0.7493 $\pm$ 0.0582& 0.9343 $\pm$ 0.0409& 0.9378 $\pm$ 0.0436& 0.9539 $\pm$ 0.0287& 0.8968 $\pm$ 0.0729 \\
  \bottomrule
  \end{tabular}}
\end{table*}

\paragraph{\edited{RSVP controls.}}
\edited{LDA predicts the current target/non-target label $y_t$ from epochs aligned to trials $t-10$, $t-1$, and $t$ (Table~\ref{tab:rsvp_shifted}). The truncated $[0,0.48)$-s window ends before the shortest observed onset interval of 482 ms. At the nominal 500-ms interval, the full $t-1$ window includes about 400 ms after $t$ begins; truncation reduces its AUC from 0.5825 to 0.5242. Filtering, baseline correction and physiological carry-over can still affect the truncated features.}

\begin{table}[htbp]
  \centering
  \color{\editcolor}
  \caption{Shifted-trial LDA controls. Entries are mean AUC with subject-bootstrap 95\% confidence intervals. All columns predict the current label $y_t$.}
  \label{tab:rsvp_shifted}
  \small
  \begin{tabular}{lccc}
    \toprule
    Feature window & $X_{t-10}$ & $X_{t-1}$ & $X_t$ \\
    \midrule
    $[-0.2,0.9]$ s & 0.4900 & 0.5825 & 0.7206 \\
     & [0.4820, 0.4984] & [0.5681, 0.5969] & [0.7005, 0.7411] \\
    $[0,0.48)$ s & 0.4987 & 0.5242 & 0.6598 \\
     & [0.4917, 0.5063] & [0.5172, 0.5311] & [0.6435, 0.6760] \\
    \bottomrule
  \end{tabular}
\end{table}

\edited{Whole-block LDA trains on three acquisition blocks and tests on the fourth. The minimum train/test onset gap is 5.699 s, exceeding the 1.1-s epoch span. AUC is $0.7225$, compared with $0.7206$ using shuffled folds. Onset-truncated, neighbor-conditioned whole-block AUC is $0.6002$ (95\% CI $[0.5859,0.6134]$), retaining only trials whose immediate neighbors are both non-target. Appendix~\ref{appendix:rsvp} gives the protocols and per-task results.}

\subsection{Application: Brain-Supervised Semantic Image Editing}
\label{sec:application}

A key contribution of our dataset is the direct pairing of each neural recording with the latent vector used to synthesize its corresponding visual stimulus.
This unique structure enables novel applications that bridge cognitive neuroscience and generative AI.
As a proof-of-concept, we demonstrate how the dataset can be used for semantic image editing supervised directly by a user's implicit neural responses, bypassing the need for explicit feedback from the user.

\begin{figure*}[h]
    \centering
    \includegraphics[width=1.0\linewidth]{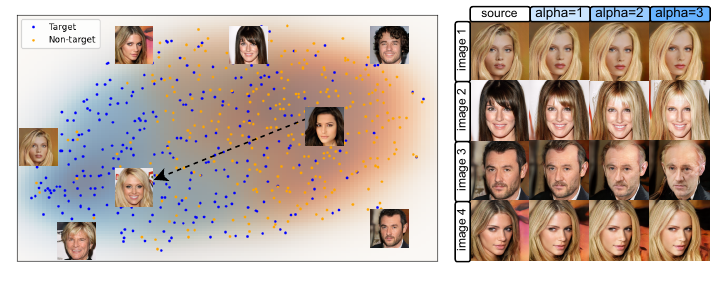}
    \vspace{-1.0cm}
    \caption{
    Demonstration of dataset utility for brain-supervised editing on the task \emph{blond}.
    \textbf{Left}: \edit{A UMAP projection of a self-supervised contrastive joint EEG-latent embedding (no label supervision); target (blue) and non-target (orange) trials cluster separately.}
    \textbf{Right}: A practical semantic editing application. A semantic direction vector is derived from a single-subject LDA model. Applying this vector with a scalar $\alpha$ progressively edits the source images to align with the subject's implicit neural representation of the target concept \emph{blond}.
    }
    \label{fig:umap}
\end{figure*}

As illustrated in Figure~\ref{fig:umap}, we can learn a user's internal concept of a semantic feature, such as "blond hair".
We train a simple LDA classifier on a single subject's data for the "blond" (target) and "dark hair" (non-target) using EEG signals.
Then a linear support vector regression (SVR) model is trained to predict the LDA's decision function from the latent vectors.
The weight vector of this SVR defines a semantic direction in the high-dimensional latent space of the generative model.
By moving a new, unseen image's latent vector along this neurally-derived direction, we can edit the image to align with that specific user's cognitive representation of "blondness", which can be a highly personalized concept.
The right panel of Figure~\ref{fig:umap} shows this process in action: Extrapolating this vector with a scalar $\alpha$ progressively modifies source images along this subject-specific semantic axis, turning dark-haired images into blond ones.

This demonstrates a powerful capability of the dataset: to move beyond discrete classification and model the continuous, subjective nature of human perception.
\edit{The left panel of Figure~\ref{fig:umap} shows a UMAP projection of a joint EEG-latent embedding learned by self-supervised contrastive alignment between EEG epochs and the latent vector that produced the corresponding stimulus, with no use of the binary semantic labels.
Even in this label-free embedding, target (blue) and non-target (orange) trials cluster separately, indicating that the dataset's paired structure carries learnable semantic information independently of the supervised editing pipeline above.}
It is important to note that the latent space is uniformly sampled from the Gaussian distribution used by the progressive GAN model, which does not have any explicit clustering structure.

\edit{\subsection{Application: Brain-Supervised Conditional Image Generation}}
\label{sec:application_gen}

\edit{The dataset also supports fully label-supervised conditional generation: an MLP brain decoder is trained to predict the binary task label from EEG, its predictions condition a 512-dimensional adversarial autoencoder (AAE)~\cite{ma2023brain} on the latent space of the same progressive GAN that produced the stimuli, and toggling the conditioning label at inference yields paired images that share source content but differ along the target semantic axis. We compare this brain-conditioned model against an \emph{explicit} variant conditioned on the manually annotated labels (empirical upper bound) and a \emph{random} variant conditioned on permuted labels (empirical lower bound). Generated pairs are evaluated by a blinded annotator with left/right order randomized; the annotator picks the image that better expresses the target attribute or marks the pair indistinguishable. Architecture, training schedule, and full annotation protocol are in Appendix~\ref{appendix:aae}.}

\edit{Figure~\ref{fig:aae_bar} reports per-task correct-selection rates (matches out of $40$ pairs, Wilson 95\,\% binomial CIs). Averaged across the eight tasks the brain-supervised model reaches $0.747$ versus $0.516$ for random and $0.825$ for the label upper bound; a two-sample $t$-test rejects brain-vs-random at $p < 10^{-15}$, while the brain-vs-explicit gap is not significant under Bonferroni correction. Qualitative paired generations (Appendix~\ref{appendix:aae:samples}) show brain and explicit conditioning flipping the target attribute while preserving non-target content; random conditioning does not.}

\begin{figure}[t]
  \centering
  \includegraphics[width=1.0\linewidth]{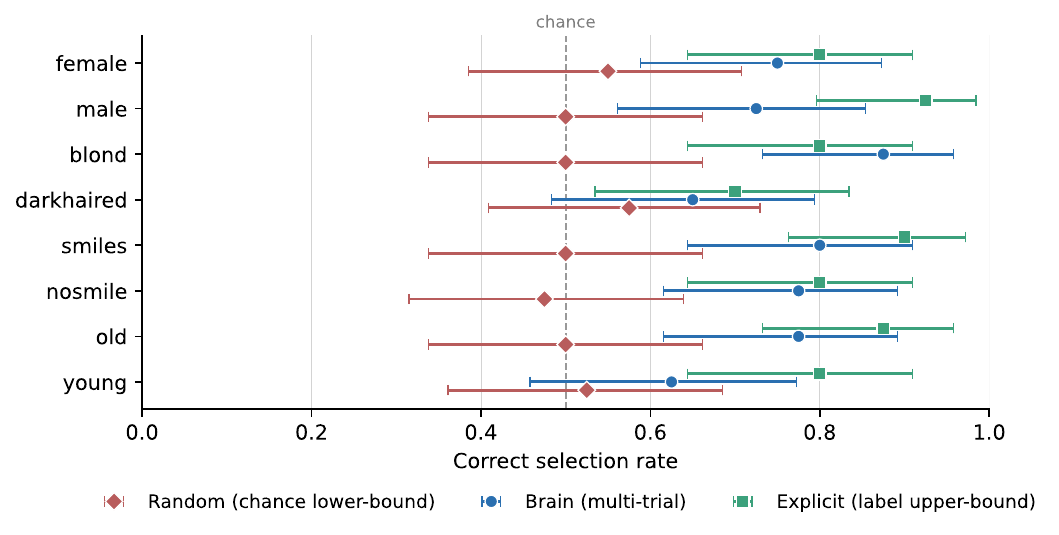}
  \edit{\caption{Per-task correct-selection rate of the AAE-generated paired images under blinded paired annotation. Markers: fraction of $40$ pairs whose target image was correctly selected; whiskers: Wilson 95\,\% binomial CIs; dashed line: chance ($0.5$). Random sits on chance everywhere; brain is well above chance on every task and approaches the explicit upper bound. Means: brain $0.747$, random $0.516$, explicit $0.825$.}\label{fig:aae_bar}}
\end{figure}

\section{Discussion}
\label{sec:discussion}

The benchmark experiments presented in Section \ref{sec:experiments} provide a quantitative baseline for our dataset, but their primary utility is in illuminating the unique computational challenges and research opportunities it enables.
We structure our discussion around three principal axes: the implications of our benchmark results, the new research frontiers this dataset opens for the computer vision community, and an honest appraisal of the dataset's limitations.

\edit{\subsection{Interpretation of Benchmark Performance}
Single-subject mean AUCs in the $0.7$--$0.75$ range, with a strong showing for linear LDA, confirm a robust decodable neural signal in every task. The gap between single-subject and cross-subject (LOSO) results, together with the high single-subject standard deviation, establishes inter-subject variability as the dataset's principal modelling challenge; the subject-adapted setting frames this as few-shot personalisation. Cross-subject voting yields substantially higher AUCs than any single-trial setting, supporting multi-subject training for foundation-style EEG models.}

\edit{\subsection{New Frontiers for Brain-Supervised Learning}
Beyond the two applications above, the paired EEG-latent structure of our dataset opens further directions: cognition-supervised visual saliency replacing explicit attention annotations\newedit{~\cite{ma2024cognition}}, cross-modal representation learning over a continuous visual latent space, and passive BCI for cognition-based image retrieval.}

\edit{\paragraph{Limitations.}
Block-design considerations and the within-task-only restriction are stated in Section~\ref{sec:block_design}. Two further limitations: (i) stimuli are exclusively CelebA-HQ-style faces and inherit that distribution\edited{, including its demographic skew}, so our dataset is best treated as a controlled-faces benchmark; (ii) the dataset is unimodal (visual), and future work could extend the paradigm to multi-modal stimuli.}

\edit{\paragraph{Conclusion.}\label{sec:conclusion}
EEGFaceSem pairs EEG with the generative latents that produced each stimulus, supporting four benchmark protocols and two applications: brain-supervised semantic image editing and brain-supervised conditional image generation. Dataset, benchmark code, and analysis scripts are released openly.}

\edit{\paragraph{Data and code.}\label{sec:availability}
EEG data, stimuli, generative latents, annotations, and benchmark code are released at OSF (\url{https://osf.io/2guk3/}) and HuggingFace (\url{https://huggingface.co/datasets/yefllower/EEGFaceSem}); a Datasheet~\cite{gebru2021datasheets} accompanies the OSF release.}

\clearpage
{\small
\expandafter\def\csname crbib@cr_grizou2025selfcalibrating\endcsname{}
\expandafter\def\csname crbib@cr_torre2024perceptual\endcsname{}
\expandafter\def\csname crbib@cr_torre2020relevance\endcsname{}
\expandafter\def\csname crbib@cr_kangassalo2020neuroadaptive\endcsname{}
\expandafter\def\csname crbib@ma2024cognition\endcsname{}
\let\originalbibitem\bibitem
\renewcommand{\bibitem}[2][]{%
  \ifcsname crbib@#2\endcsname\color{\reviewcolor}\else\color{black}\fi
  \originalbibitem[#1]{#2}}
\bibliographystyle{plainnat}
\bibliography{main,camera_ready_additions}
}

\newpage
\appendix

\edit{\section{AAE Generative Pipeline: Architecture, Training, Annotation}}
\label{appendix:aae}

\edit{This appendix expands Section~\ref{sec:application} with the architecture, training schedule, annotation protocol, and per-task numerical results of the brain-supervised generative pipeline introduced in prior work~\cite{ma2023brain}.
The numbers and figures reported here are recomputed from the released annotation log using the scripts bundled with the dataset code, rather than reproduced from the original publication.}

\edit{\subsection{Brain decoder}}
\label{appendix:aae:decoder}

\edit{For each participant and each semantic task an MLP is trained that maps a single EEG epoch to a binary target/non-target probability.
The input is a flattened $32 \times W$ tensor covering $-200$ to $+900\,\mathrm{ms}$ relative to stimulus onset (the same epoch window released with the dataset).
The network has two fully connected hidden layers of width 2{,}048 with leaky-ReLU activation and dropout 0.2, followed by a sigmoid output.
Training uses Adam with learning rate $10^{-4}$ and follows the schedule documented in the released benchmark code.
For each held-out stimulus a brain-derived label is produced by aggregating the per-subject predictions of all participants who viewed that stimulus and taking the median posterior; this matches the cross-subject voting protocol used in Section~\ref{sec:tasks_and_results}.}

\edit{\subsection{Adversarial autoencoder}}
\label{appendix:aae:gen}

\edit{A conditional AAE is trained on the 512-dimensional latent vectors used to render the stimulus images.
The encoder maps a latent vector to a 128-dimensional code through two fully connected layers of width 1{,}000 with ReLU; the decoder mirrors this and concatenates the code with the binary conditioning label before reconstructing the original 512-dimensional latent.
A linear-output MLP critic with two width-1{,}000 ReLU layers regularizes the code distribution toward a standard normal via a Wasserstein objective.
Training alternates a reconstruction phase and a regularization phase with an Adam optimizer, learning rate $10^{-4}$, and balancing weight $\lambda = 0.9$ for all tasks.
The conditioning label is one of three possibilities depending on the experimental cell:
the manually annotated label (the explicit model), the brain-derived aggregated label (the brain model), or a randomly permuted label (the random control).
Running the full pipeline (decoder, AAE, generator) for all $30$ participants $\times$ $8$ tasks $\times$ $3$ models takes roughly ten GPU-hours on a single consumer GPU.}

\edit{\subsection{Annotation protocol}}
\label{appendix:aae:anno}

\edit{For every (task, model) cell, $20$ source latent vectors are sampled from a held-out partition of the dataset; each source produces two images by toggling the conditioning label, yielding $40$ images per cell.
The annotator views the two images side-by-side with left/right order independently randomized for each pair and chooses (i) the image that more saliently expresses the target attribute, (ii) the other image, or (iii) ``same'' when the difference is too subtle to judge.
We map these into a per-pair score: $1$ if the chosen image matches the target label, $0$ otherwise, with a separate ``same'' indicator.
The headline metric reported in Section~\ref{sec:application} is the correct-selection rate, defined as the count of pairs in (i) divided by $40$.
A soft variant $\hat r = r + \tfrac{1}{2} f_{\mathrm{same}}$ that credits half a point for ``same'' pairs (capturing the equal-target-density case in which an indistinguishable pair contributes to both labels in expectation) is reported alongside in Table~\ref{tab:aae_full}.}

\edit{To verify reliability, an independent annotator who did not participate in the EEG experiment re-labelled a random subset of $480$ pairs.
Cohen's kappa between the two annotators on this subset is $0.76$, indicating substantial agreement.}

\clearpage
\edit{\subsection{Per-task results}}
\label{appendix:aae:results}

\begin{table}[!htbp]
\centering
\edit{\caption{Full per-task results of the AAE pipeline at training step $1{,}000$. $r$ is the correct-selection rate (correct out of $40$); $\hat r$ is the soft variant that credits ``same'' pairs as half a point.}\label{tab:aae_full}}
\begin{tabular}{l|cc|cc|cc}
\toprule
& \multicolumn{2}{c|}{Brain} & \multicolumn{2}{c|}{Explicit} & \multicolumn{2}{c}{Random} \\
Task & $r$ & $\hat r$ & $r$ & $\hat r$ & $r$ & $\hat r$ \\
\midrule
female      & 0.750 & 0.90 & 0.800 & 0.90 & 0.550 & 0.62 \\
male        & 0.725 & 0.78 & 0.925 & 0.93 & 0.500 & 0.53 \\
blond       & 0.875 & 0.97 & 0.800 & 0.97 & 0.500 & 0.50 \\
darkhaired  & 0.650 & 0.72 & 0.700 & 0.80 & 0.575 & 0.57 \\
smiles      & 0.800 & 0.93 & 0.900 & 0.95 & 0.500 & 0.50 \\
nosmile     & 0.775 & 0.78 & 0.800 & 0.90 & 0.475 & 0.42 \\
old         & 0.775 & 0.88 & 0.875 & 0.95 & 0.500 & 0.45 \\
young       & 0.625 & 0.88 & 0.800 & 0.95 & 0.525 & 0.65 \\
\midrule
Mean        & \textbf{0.747} & 0.853 & \textbf{0.825} & 0.919 & \textbf{0.516} & 0.531 \\
\bottomrule
\end{tabular}
\end{table}

\edit{The brain model exceeds the random control on all eight tasks (binomial test, $p < 0.05$ on every task; $p < 10^{-15}$ when pooled across tasks).
The brain-vs-explicit gap is small (mean difference $0.078$ in $r$, $0.066$ in $\hat r$) and not significant under a Bonferroni-corrected $t$-test across tasks.
This pattern shows that the brain-derived labels carry semantic information of comparable, though slightly lower, quality to the manually annotated labels.}

\edit{The pipeline summarised here is the one introduced in~\cite{ma2023brain};
the present paper releases the underlying EEG dataset, re-tabulates the annotation log, and regenerates the figures from the released benchmark code.}

\clearpage
\edit{\subsection{Qualitative paired generations}}
\label{appendix:aae:samples}

\begin{figure}[!htbp]
  \centering
  \includegraphics[width=0.85\linewidth]{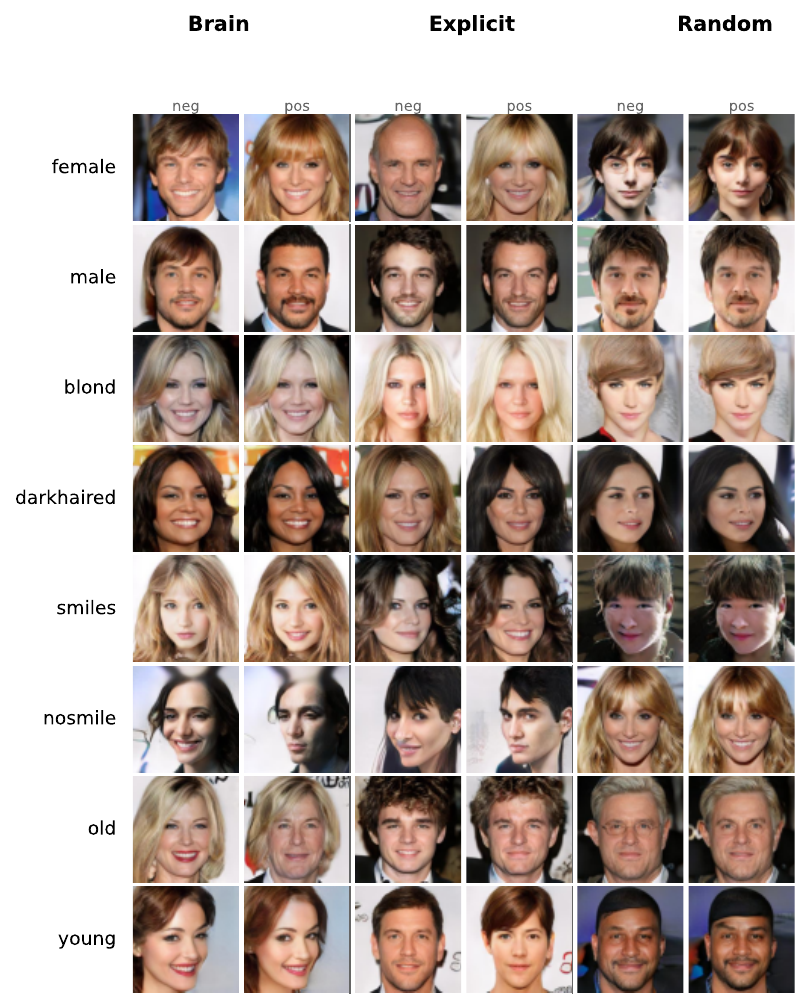}
  \edit{\caption{Paired generations under brain, explicit-label, and random-label conditioning, one example pair per (task, model) at training step $1{,}000$. The brain and explicit columns flip the target attribute (e.g.\ blond/dark, smiling/not) while preserving identity-level features; the random column does not.}\label{fig:aae_grid}}
\end{figure}

\clearpage
\edit{\section{External Attribute Audit}}
\label{appendix:attribute_audit}

\edit{We probe each released stimulus with CLIP ViT-L/14~\cite{radford2021clip} in a zero-shot configuration, using positive/negative prompt pairs for the four target attributes (gender, smile, hair colour, age). Table~\ref{tab:attribute_audit_separation} reports per-contrast separation between the screened positive and negative groups: every contrast is recovered with AUC $\geq 0.816$, with \edited{two} out of four above $0.99$. Table~\ref{tab:attribute_audit_delta_phi} reports $\Delta\Phi$, the difference in pairwise phi coefficients between the screened stimuli and a reference batch of $N{=}5{,}000$ random latent samples drawn from the same generator. The largest off-diagonal entry has $|\Delta\Phi|=0.015$, indicating that the screening does not introduce additional pairwise attribute correlations beyond those already present in the unfiltered generator output. This is an external check that the released stimuli express the intended semantic contrasts and do not collapse into trivially co-occurring categories.}

\begin{table}[!htbp]
  \centering
  \small
  \edit{\caption{Residual pairwise attribute association introduced by screening: $\Delta\Phi = \Phi(\text{screened}) - \Phi(\text{random})$. Values close to zero indicate that the screening does not introduce additional attribute confounds beyond those already present in the generator's output distribution.}\label{tab:attribute_audit_delta_phi}}
  \begin{tabular}{lcccc}
    \toprule
    & Gender & Smile & Blond & Young \\
    \midrule
    Gender & -- & $-0.007$ & $-0.011$ & $0.010$ \\
    Smile & $-0.007$ & -- & $0.005$ & $0.010$ \\
    Blond & $-0.011$ & $0.005$ & -- & $-0.015$ \\
    Young & $0.010$ & $0.010$ & $-0.015$ & -- \\
    \bottomrule
  \end{tabular}
\end{table}

\clearpage
\section{\edited{RSVP timing and whole-block controls}}
\label{appendix:rsvp}

\edited{For Table~\ref{tab:rsvp_shifted}, offsets follow raw event order within the same task and block; both epochs must survive preprocessing. Offsets $t-10$, $t-1$, and $t$ retain 45,882, 54,127, and 59,348 pairs. Five shuffled folds (seed 42) assign pairs by the current trial, separately for each participant and task.}

\edited{Features are flattened EEG, scaled by 1,000 as in the loader, followed by a training-fitted StandardScaler and default scikit-learn LinearDiscriminantAnalysis. Full and truncated windows contain 1,101 and 480 samples per channel. Fold AUCs are averaged within participant/task and then equally across 30 participants and eight tasks.}

\edited{For whole-block evaluation, matching to raw events excludes three epochs and corrects one task-code anomaly, leaving 59,345 epochs across 960 folds. The largest task-mean AUC difference from shuffled folds is 0.0087. This rerun covers LDA only. Neighbor conditioning excludes block edges and retains 30,438 trials (16,774 targets; 55.1\%), using raw labels to identify neighbors.}

\begin{table}[htbp]
  \centering
  \color{\editcolor}
  \caption{Whole-block LDA AUC. ``Neighbors non-target'' also uses $[0,0.48)$-s current-trial features. Entries are task means across 30 participant fold-means. Mean-row SD is descriptive variation across eight task means.}
  \label{tab:rsvp_grouped}
  \small
  \begin{tabular}{lccc}
    \toprule
    Task & Full window & Truncated window & Neighbors non-target \\
    \midrule
    Female & 0.7888 & 0.7194 & 0.6359 \\
    Male & 0.7296 & 0.6672 & 0.6225 \\
    Blond & 0.7744 & 0.7092 & 0.6444 \\
    Darkhaired & 0.7395 & 0.6920 & 0.6335 \\
    Smiles & 0.7306 & 0.6660 & 0.5969 \\
    Nosmile & 0.7076 & 0.6302 & 0.5704 \\
    Old & 0.6583 & 0.5959 & 0.5420 \\
    Young & 0.6512 & 0.6022 & 0.5559 \\
    \midrule
    Mean $\pm$ SD & $0.7225\pm0.0460$ & $0.6603\pm0.0439$ & $0.6002\pm0.0372$ \\
    \bottomrule
  \end{tabular}
\end{table}

\edited{Confidence intervals use 10,000 joint participant-bootstrap resamples (seed 20260728), preserving tasks within participants and averaging folds before resampling. The truncated whole-block AUC without neighbor conditioning is 0.6603 (95\% CI [0.6441,0.6758]).}

\end{document}

%% file: preamble.tex
\newcommand{\edit}[1]{#1}